\documentclass[letterpaper, 10 pt, conference]{ieeeconf}  % Comment this line out if you need a4paper

\IEEEoverridecommandlockouts                              % This command is only needed if 
\usepackage{graphics} % for pdf, bitmapped graphics files
\usepackage{epsfig} % for postscript graphics files
\usepackage{multirow}
\usepackage{amsmath} % assumes amsmath package installed
\usepackage{amssymb}  % assumes amsmath package installed
\usepackage{graphicx}
\usepackage{subfigure}
\usepackage{booktabs}

\title{\LARGE \bf
EVIT: Event-based Visual-Inertial Tracking in Semi-Dense Maps\\Using Windowed Nonlinear Optimization
}

\author{Runze Yuan, Tao Liu, Zijia Dai, Yi-Fan Zuo, and Laurent Kneip% <-this % stops a space
\thanks{Mobile Perception Lab, ShanghaiTech University}% <-this % stops a space
}

\begin{document}

\maketitle
\thispagestyle{empty}
\pagestyle{empty}

%%%%%%%%%%%%%%%%%%%%%%%%%%%%%%%%%%%%%%%%%%%%%%%%%%%%%%%%%%%%%%%%%%%%%%%%%%%%%%%%
\begin{abstract}

Event cameras are an interesting visual exteroceptive sensor that reacts to brightness changes rather than integrating absolute image intensities. Owing to this design, the sensor exhibits strong performance in situations of challenging dynamics and illumination conditions. While event-based simultaneous tracking and mapping remains a challenging problem, a number of recent works have pointed out the sensor's suitability for prior map-based tracking. By making use of cross-modal registration paradigms, the camera's ego-motion can be tracked across a large spectrum of illumination and dynamics conditions on top of accurate maps that have been created a priori by more traditional sensors. The present paper follows up on a recently introduced event-based geometric semi-dense tracking paradigm, and proposes the addition of inertial signals in order to robustify the estimation. More specifically, the added signals provide strong cues for pose initialization as well as regularization during windowed, multi-frame tracking. As a result, the proposed framework achieves increased performance under challenging illumination conditions as well as a reduction of the rate at which intermediate event representations need to be registered in order to maintain stable tracking across highly dynamic sequences. Our evaluation focuses on a diverse set of real world sequences and comprises a comparison of our proposed method against a purely event-based alternative running at different rates.

\end{abstract}

%%%%%%%%%%%%%%%%%%%%%%%%%%%%%%%%%%%%%%%%%%%%%%%%%%%%%%%%%%%%%%%%%%%%%%%%%%%%%%%%
\section{INTRODUCTION}

Event cameras have recently become popular in the vision and robotics community but remain less explored than their traditional camera correspondents. Unlike the latter, event cameras---also called Dynamic Vision Sensors (DVS)---react to brightness changes rather than integrating absolute image intensities. Each pixel acts asynchronously and fires an either positive or negative event whenever the perceived brightness level augments or decrements by a certain threshold amount. Each event comes with a high resolution time-stamp in the order of micro-seconds, and the camera has high dynamic range (120dB compared to 60dB for regular cameras). Owing to these properties, event cameras do not suffer from traditional blur effects and lend themselves to strong performance in challenging illumination and dynamics conditions. Often referred to as a silicon retina~\cite{mahowald91}, this power-efficient bio-inspired sensor therefore has evolved into a promising choice for motion sensing in a variety of contexts, including eye tracking~\cite{Angelopoulos22}, hand tracking~\cite{rudnev20}, human body tracking~\cite{xu20,zhang23}, and---as targeted in this work---ego-motion tracking.
\begin{figure}[h]
	\centering
        \includegraphics[width=0.90\columnwidth]{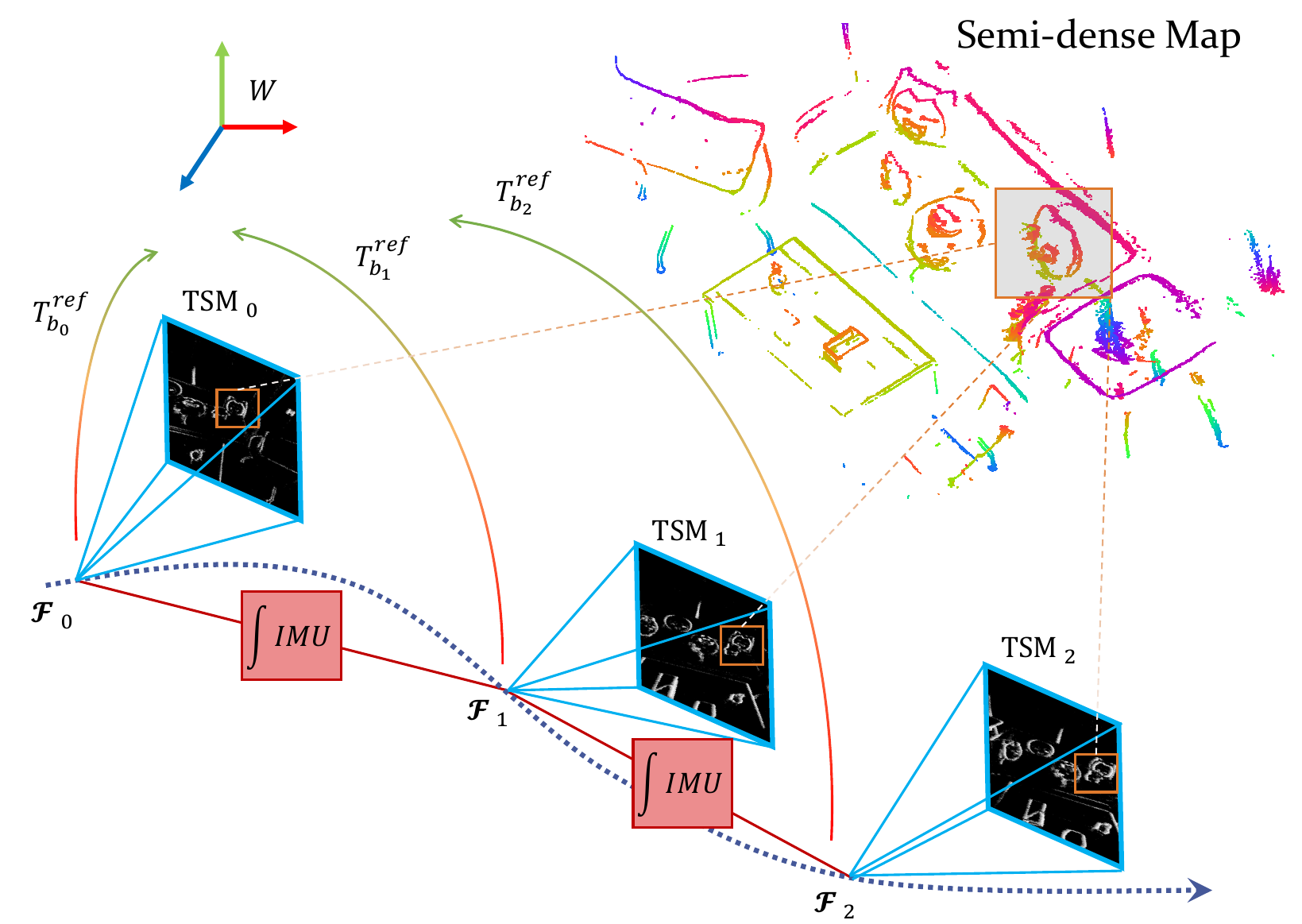}
	\caption{Illustration of EVIT. Rather than registering only single time-surface maps with respect to a semi-dense point cloud, we propose to do windowed joint registration of multiple adjacent TSMs, which improves registration stability. The added IMU integration terms form the connections between adjacent keyframes, thereby creating a virtual multi-camera rig with elastic connections.}
	\label{fig:font_pic}
\end{figure}

The traditional solution to vision-based ego-motion tracking is given by the Simultaneous Localization And Mapping (SLAM) paradigm~\cite{cadena2016past}, and has already been proposed for event cameras~\cite{kim2016real,rebecq2016evo}. However, unlike for regular cameras, the measurements returned by an event camera are no longer just a function of pose and scene geometry or appearance, but also the instantaneous relative dynamics of the camera. Combined with the unusual and noisy nature of events, the solution to the full SLAM problem with DVS sensors remains an ongoing challenge.

In this paper, we address the question how we can solve the ego-motion estimation problem under the often practical assumption of an existing prior map generated by a different sensor? If the mapping problem is taken out of the equation, the target simply becomes prior map-based tracking. We follow up on the recently proposed strategy of abstracting the map of the environment in the form of a semi-dense point cloud~\cite{zuo2024cross}. The semi-dense map can be generated from regular imagery, and each point in the representation corresponds to a location in space around which we commonly observe high appearance gradient (i.e. structural edges and discontinuities, or textural appearance boundaries). Based on the assumption that the generated events pre-dominantly react to such high-gradient regions, the current location of the event camera can be optimized by aligning edges in the surface of active events with the reprojected semi-dense point cloud. While this semi-dense cross-modal registration paradigm has already been proposed in prior art~\cite{zuo2022devo,zuo2024cross}, challenges remain in case of high dynamics or ambiguous event distributions.

We propose to tackle this issue by fusion with an Inertial Measurement Unit (IMU). This achieves the following contributions and advancements:
\begin{itemize}
  \item Rather than doing single frame registration, the addition of an IMU permits the joint registration of multiple adjacent frames in the form of a windowed tracking problem. The states corresponding to the individual times of each registered frame are lifted to include first-order translational velocity and IMU bias terms, thereby enabling the use of IMU pre-integration terms for a pair-wise regularization of adjacent frames. Sets of multiple adjacent frames are thereby considered as a virtual multi-camera rig with elastic connections. The joint registration of multiple frames helps in avoiding unobservable directions of displacement caused by ambiguous event distributions in individual frames (e.g. events distributed along multiple parallel lines).
  \item IMU pre-integration terms may again be used in order to create improved initialization of new frames. Especially in situations of jerky motion with sudden abrupt changes in the direction of motion, the inertial signal-based prediction strongly outperforms vision-only alternatives such as constant-velocity motion models.
  \item Owing to poor prediction abilities, handling situations of high acceleration in a vision only setting typically requires a high temporal density of tracked frames. In turn, the superior prediction ability offered by IMU signals enables a substantial reduction of the frame density during windowed optimization. In fact, we propose an adaptive mechanism that places frames in dependence of the number of occurred events.
\end{itemize}
We test our new framework---Event-based Visual-Inertial Tracking (EVIT)---on publicly available benchmark sequences covering a variety of challenges such as normal motion, changing illumination, and high dynamics. As demonstrated, the addition of an IMU strongly supports tracking performance with respect to the vision-only alternative. We also plan to release the complete framework code later to enable re-usability.

\section{RELATED WORK}

A good survey paper on event-based vision is given by Gallego et al.~\cite{gallego2019event}. The addressed topics in the literature cover many aspects such as human motion tracking~\cite{xu20,zhang23}, hand tracking~\cite{rudnev20}, and eye ball tracking~\cite{Angelopoulos22}. The current literature review focusses on SLAM and motion tracking with only event cameras or with added inertial readings.

Pure approaches to monocular event camera SLAM have been proposed by Kim et al.~\cite{kim2016real} and Rebecq et al.~\cite{rebecq2016evo}. The former is a filter-based method reconstructing photometric depth maps, while the latter is a geometric approach that aims at the reconstruction of a volumetric event density field~\cite{rebecq2016emvs}. In an effort to increase the performance of event camera-based SLAM, more recent methods have investigated the use of line features~\cite{chamorro2022event,peng21b,gao23}. The latter two methods are limited to local dynamics estimation, and all methods are restricted in that they require a sufficient number of line features. The inclusion of an IMU into the incremental tracking and motion framework is proposed by Zhu et al.~\cite{zihao2017event}, Rebecq et al.~\cite{rebecq2017real}, Le Gentil et al.~\cite{le2020idol}, and Xu et al.~\cite{xu23tro}. In particular, the latter two methods are again based on line features. In general, a truly robust continuous tracking and mapping framework that employs only a single event camera as an exteroceptive sensor remains a challenging problem. Vidal et al.~\cite{vidal2018ultimate} therefore investigate the combination with a normal camera, and Zhou et al.~\cite{zhou2021event} present a stereo event camera alternative.

A different line of research that is more related to our work aims at pure tracking based on known structure priors. Early approaches make use of fiducial markers~\cite{censi13,chen20} or known targets with distinctive texture~\cite{mueggler14,bertrand20,chamorro20,valeiras16}. Of particular interest are the works of Gallego et al.~\cite{gallego18} and Bryner et al.~\cite{Bryner19icra}, who propose direct tracking of event cameras from photometric depth maps. However, their approaches are computationally demanding and fail to employ a geometrically optimal objective.

The most highly related works to ours are all based on the idea of using a semi-dense map representation. The latter have originally been introduced by Engel et al.~\cite{engel2013semi} through their SDVO framework. Later on, Kneip et al.~\cite{kneip2015sdicp}, Kuse and Shaojie~\cite{kuse2016robust}, and Zhou et al.~\cite{zhou2018canny} have introduced improved geometric semi-dense registration paradigms for normal cameras. The latter work in particular has formed the foundation for the first geometric semi-dense tracking paradigms for event cameras. Zuo et al.~\cite{zuo2022devo} propose DEVO, an event-based incremental motion estimation framework that makes use of an additional depth channel. Finally, Zuo et al.~\cite{zuo2024cross} propose Canny-EVT, which applies a similar strategy to global tracking of a single event camera based on a semi-dense map. The latter work in particular forms the reference implementation against which we compare our contribution.

\section{METHOD}
\begin{figure*}[t]
    \centering
    \includegraphics[width=\textwidth]{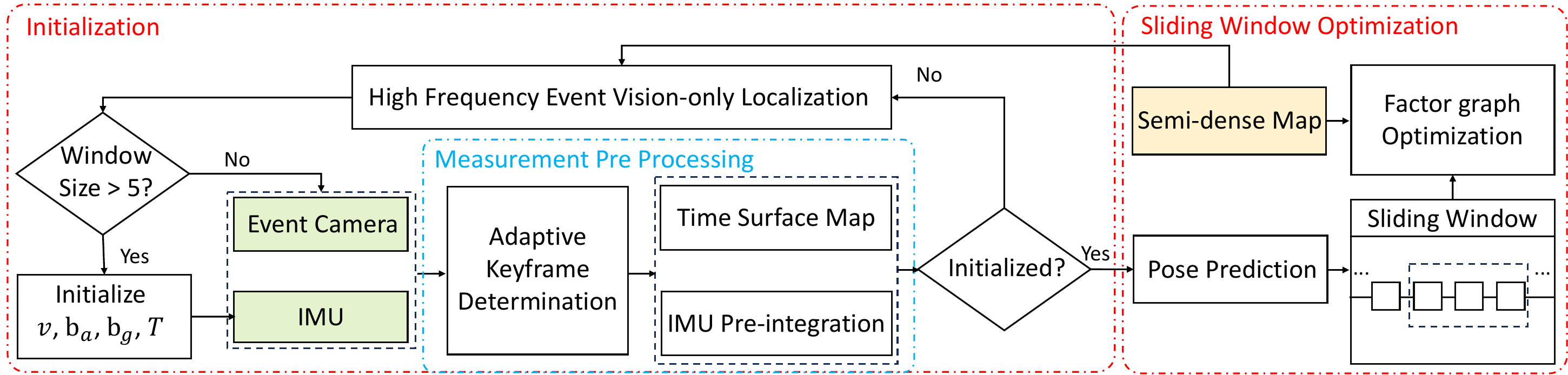}
    \caption{Block diagram of the full event-based visual-inertial tracking pipeline. The system takes stream of events and IMU measurements (colored block) as input and tracks against the reconstructed semi-dense map. The measurement processing module (Section \ref{subsec:measurement processing}) dynamically choose keyframes and process raw data stream into usable single frame observations. The initialization (Section \ref{subsec:init}) module utilizes high frequency event localization results to provide bootstrapping states for subsequent prediction and optimization. The optimization module (Section \ref{subsec:optimization}) tightly fuses IMU pre-integration measurements and \textbf{TSM} representations to achieve accurate state estimation. }
    \label{fig:pipeline}
\end{figure*}

In this section, we present our newly proposed event-inertial, semi-dense point cloud-based tracking system EVIT. The framework comprises an efficient adaptive measurement preprocessing module, a loosely coupled bootstrapping scheme, and a tightly coupled back-end optimization for the event camera-IMU setup.
The structure of the proposed framework is shown in Fig.~\ref{fig:pipeline}, in which we highlight the essential modules by dashed lines. 

Let us define notations used in this paper. The ultimate goal for this framework is to estimate the body frame pose $\mathbf{T}^w_b = [\mathbf{R}^w_b, \mathbf{p}^w_b] \in \text{SE}(3)$ (transformation from the body frame to the world frame) of the event camera and IMU rig. For practical reasons, we set the IMU frame equal to the body frame, denoted by b. The constant extrinsic transformation $\mathbf{T}^c_b$ between IMU and event camera is assumed to be well calibrated. Visual-inertial tracking introduces additional variables to the state estimation, leading to the expanded state vector
\begin{align}
    \boldsymbol{{\mathcal{S}}}_i = [\boldsymbol{\theta}_i, \mathbf{p}_i,  \mathbf{v}_i, \mathbf{b}_{\alpha_i}, \mathbf{b}_{\omega_i}],
\end{align}
where $\boldsymbol{\theta}_i$ is a minimal rotation representation, $\mathbf{p}_i$ is the position of the body frame, $\mathbf{v}_i$ is the velocity in the world reference frame, and $ \mathbf{b}_{\alpha_i}$ and $\mathbf{b}_{\omega_i}$ are the accelerometer and gyroscope biases, respectively.

\subsection{System overview}
We start by introducing the entire pipeline and describing the functionality of each core module in our framework. The framework starts with an offline-constructed, inertial-aligned semi-dense point cloud and takes as input a stream of events and IMU measurements from a calibrated event-inertial sensor setup. A specially designed data collection module is responsible for the asynchronous collection of these two types of data thereby achieving maximum effectiveness. 
The data collection module is followed by a measurement processing module that takes the quasi-continuous input stream of data and converts it into compact single frame observations (i.e. time surface maps and IMU pre-integration terms). The generated frames are henceforth referred to as keyframes. Details on the intermediate, frame-based event representation are provided in Section \ref{subsec:measurement processing}. Note that the rate at which keyframes are generated is adaptive and depending on the actual number of events that have elapsed since the last keyframe generation. The generation of IMU pre-integration terms adjusts to the asynchronous keyframe rate. The proposed dynamic, adaptive frame generation strategy simultaneously satisfies two objectives. First, it avoids the generation of redundant keyframes when no actual change or camera motion has happened. Second, it avoids a lack of sufficient observations under fast displacements. 

During the initialization phase (cf. Section \ref{subsec:init}), we perform high-frequency event-only single keyframe localization to retrieve initial body frame poses. When the displacement is sufficient, a closed-form visual-inertial solver is executed in order to give coarse initial values for IMU states to be estimated, meaning the camera velocity and IMU biases.
After successful bootstrapping, the tracking module (see Section \ref{subsec:optimization}) starts to predict poses for new keyframes using a second-order motion model (i.e. using IMU forward integration) and activates a tightly coupled event-inertial back-end optimizer employing a sliding window over a fixed number of recent keyframes. Rather than only optimizing keyframe poses, this factor graph optimization module optimizes the full above-mentioned inertial states including camera velocity and biases for each keyframe. The addition of inter-frame pre-integration terms adds regularization constraints between adjacent keyframes, and thereby transforms the registered set of keyframes into a virtual multi-camera rig with elastic extrinsic connections.

\subsection{Measurement preprocessing}
\label{subsec:measurement processing}

In traditional visual-inertial tracking frameworks, it is common to select keyframes based on image rate given that conventional cameras are synchronously triggered and usually operate at frequencies lower than IMUs. Considering a single event provides limited information, for event cameras we are equally inclined to preprocess events and somehow group them in batches that are triggered within a specific time interval so they can be jointly aligned against the map. In analogy to the work of Zuo et al.~\cite{zuo2024cross}, we propose to summarize the contribution of entire intervals of events in snap-shots of the Time Surface Map (TSM), which can then be used to extract edges and perform individual frame alignment against the reprojected semi-dense map. The question is at what rate such TSM-based keyframes should be generated.

\subsubsection{Adaptive keyframe determination}
An event camera measures motion with respect to the scene, and the rate at which events are generated is related to the actual amount of displacement. 
The fixed frequency keyframe selection scheme proposed in previous works \cite{zhou2021event} therefore is sub-ideal and would lead to inferior optimization performance. It would either lead to redundant frames if little motion occurs, or insufficient frame density and large inter-keyframe baselines under fast displacements.
Considering the IMU, the pre-integration also requires sufficient readings in order to ensure noise cancellation effects and good Signal-to-Noise Ratio (SNR).
We therefore propose an adaptive keyframe generation scheme that buffers events and inertial readings and---only when a minimum number of both  events and IMU readings has been surpassed---outputs the accumulated series of observations to the frame builder. The latter then processes the raw data into TSMs and IMU pre-integration terms. The thresholds for the minimum number of events and IMU observations are denoted $n_{event} ~ \text{and} ~ n_{imu}$, respectively. The time stamp of the keyframe is determined by the last IMU measurement.

\subsubsection{Event representation}
As mentioned, we rely on snap-shots of the Time Surface Map (TSM) for semi-dense registration. The latter is efficient to update and does not involve costly feature extractors. It is defined as follows. Let us assume a stream of $N$ events denoted as $e_i=\{\mathbf{x_i},t_i,p_i\},i\in\mathbb{N}$, which includes location $\mathbf{x_i}=\{u_{i},v_{i}\}$, timestamp $t_i$ and polarity $p_{i}\in\{-1,1\}$.
% They contain richer spatiotemporal information compared to traditional images due to their asynchronous generation. Thus, it's crucial to represent these event streams efficiently for the implementation of subsequent algorithms.
%
A time surface map is a 2D map in which the value at a pixel $\mathbf{x}$ is defined by the most recent event occurring at that location. It is notably depending on an exponential decay kernel defined as 
% A widely employed representation in event-based vision is the \textbf{TSM} (Time Surface Map), which assigns a value to each pixel at arbitrary time and utilizes an exponential decay kernel, 
\begin{align}
\mathcal{T}(\mathbf{x},t)\doteq \exp\left(-\frac{t-t_{last}(\mathbf{x})}{\delta}\right)
\end{align}
where $\delta$ is a constant decay rate parameter typically set to a small value. $t_{\mathrm{last}}(\mathbf{x})$ is the timestamp corresponding to the most recent triggered event at $\mathbf{x}$. $t$ is a given time stamp for which the TSM is constructed. Note that in the context of our work, the TSM can be regarded as the historical trajectory of scene edges moving in the image plane, and sharp contours remain identifiable at the reprojection location of the currently observed edges in the scene. It is this phenomenon that encourages a direct use of TSMs towards semi-dense registration.

For practical use, we usually use the inverse (or negated) TSM given by 
\begin{align*}
    \overline{{\mathcal{T}}}(\mathbf{x},t) = 1 - \mathcal{T}(\mathbf{x},t).
\end{align*}
It has the advantage that the locations corresponding to the reprojection of the currently observed edges now present the lowest values in the map. As a result, the negated TSM can be directly used as a cost field against which the reprojected semi-dense map can be registered in nonlinear optimization.
The relationship between an actual distance field and the negated TSM is introduced in \cite{zhou2021event}. We may readily perform nonlinear optimization on this geometric field.

For the sake of efficient measurement preprocessing, we maintain a global surface of active events (SAE) \cite{benosman2013event} which stores the most recent event time stamp at each pixel. We also map the pixel values in the TSM from $[0,1]$ to $[0,255]$ for easy visualization and stable numerical computation. We make use of the truncated TSM in which all values below a given threshold $\delta$ are set to 0. Finally, we apply a Gaussian kernel on the TSM to further smoothify the field, reduce ripple effects, and reduce the influence of outlier events.

% \textbf{TSM} employs a time-focused methodology to derive spatio-temporal features from the dynamic aspects of a visual scene, which are acquired asynchronously\cite{hots2017}. 

\subsubsection{Imu pre-integration}
The IMU provides measurements for acceleration $\boldsymbol{\alpha}_b$ and angular velocity $\boldsymbol{\omega}_b$ of the sensor at a high frequent rate.
%For the purpose of the practical implementation, it is common to set body reference frame equal to the IMU frame, denoted with \textbf{b}.
%
We follow the IMU sensor model and continuous-time quaternion-based derivation of IMU pre-integration terms introduced in previous work \cite{qin2018vins} \cite{shen2015tightly}.
The raw IMU measurements $\hat{\boldsymbol{\alpha}}_t$, $\hat{\boldsymbol{\omega}}_t$ are given by
\begin{align*}
    &\hat{\boldsymbol{\alpha}}_t = \boldsymbol{\alpha}_t + \mathbf{b}_{\alpha_t} + \mathbf{R}^t_w\mathbf{g}^w + \mathbf{n}_{\alpha} \\
    &\hat{\boldsymbol{\omega}}_t = \boldsymbol{\omega}_t + \mathbf{b}_{\omega_t} + \mathbf{n}_{\omega}
\end{align*}
where noise is modeled as Gaussian white noise, i.e. $\mathbf{n}_{\alpha}\sim\mathcal{N}(0, \boldsymbol{\sigma}^2_{\alpha}), \mathbf{n}_{\omega}\sim\mathcal{N}(0, \boldsymbol{\sigma}^2_{\omega})$, and biases are modeled as a random walk. 

The pre-integration terms between consecutive frames are given by
\begin{align*}
    & \boldsymbol{\mathcal{\alpha}}^{b_k}_{b_{k+1}} = \iint_{t\in[t_k, t_{k+1}]} \mathbf{R}^{b_k}_t(\hat{\boldsymbol{\alpha}}_t - \mathbf{b}_{\alpha_t} -  \mathbf{n}_{\alpha})dt^2 \\
    & \boldsymbol{\mathcal{\beta}}^{b_k}_{b_{k+1}} = \int_{t\in[t_k, t_{k+1}]} \mathbf{R}^{b_k}_t(\hat{\boldsymbol{\alpha}}_t - \mathbf{b}_{\alpha_t} -  \mathbf{n}_{\alpha})dt \\
    & \boldsymbol{\mathcal{\gamma}}^{b_k}_{b_{k+1}} = \int_{t\in[t_k, t_{k+1}]}\frac{1}{2} \boldsymbol{\Omega}(\hat{\boldsymbol{\omega}} - \mathbf{b}_{\omega_t} - \mathbf{n}_{\omega})\boldsymbol{\mathcal{\gamma}}^{b_k}_tdt 
\end{align*}
where 
\begin{equation*}
    \boldsymbol{\Omega}(\boldsymbol{\omega}) = \begin{bmatrix}
        -\lfloor \boldsymbol{\omega} \rfloor_{\times} & \boldsymbol{\omega} \\
        -\boldsymbol{\omega}^T & 0
    \end{bmatrix}, 
    \lfloor \boldsymbol{\omega} \rfloor_{\times} = \begin{bmatrix}
        0 & -\omega_z & \omega_y \\
        \omega_x & 0 & -\omega_x \\
        -\omega_y & \omega_x & 0
    \end{bmatrix}
\end{equation*}
$\boldsymbol{\mathcal{\alpha}}^{b_k}_{b_{k+1}}, \boldsymbol{\mathcal{\beta}}^{b_k}_{b_{k+1}},\text{ and } \boldsymbol{\mathcal{\gamma}}^{b_k}_{b_{k+1}}$ represent the pre-integrated position, velocity and rotation measurements. A covariance matrix $\boldsymbol{\Sigma}_{b_k,b_{k+1}}$ is also calculated during the pre-integration process.
In our practical implementation, we use first-order integration for processing discrete time IMU measurements. 

\begin{figure*}[t]
    \centering
    \includegraphics[width=\textwidth]{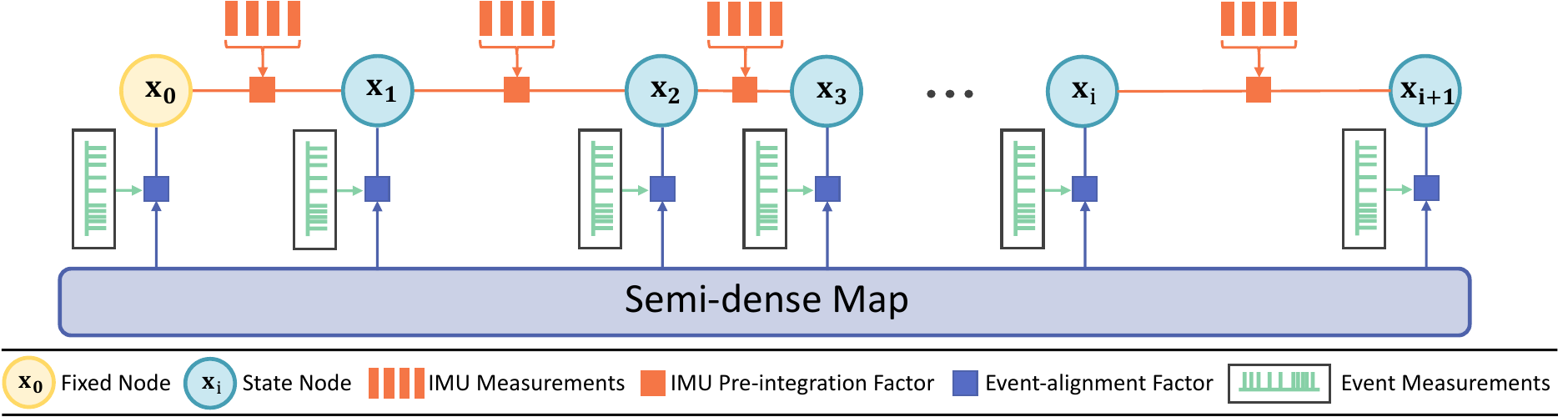}
    \caption{Factor graph representation for our sliding window optimization. The graph fuses observations from the event camera, the IMU, and the semi-dense map. Two types of factors are introduced to construct the factor graph: (a) IMU pre-integration factors, (b) event alignment factors using TSMs. The formulation of these factors are discussed in Sections \ref{subsec:init} and \ref{subsec:optimization}. We fix the first node to maintain consistency with matured nodes that left the window.}
    \label{fig:factor_graph}
\end{figure*}

\subsection{Initialization}
\label{subsec:init}
Tightly coupled visual-inertial optimization-based tracking usually needs good initial values for states to achieve convergence. The goal of the initialization module is to align vision-based and IMU measurements to calibrate the bias parameters of the IMU and provide initial states. 
Inspired by \cite{qin2018vins}, we employ a light-weight loosely coupled initialization method in our localization framework. In contrast to traditional VIO frameworks, the states to be bootstrapped in our localization system are fewer and only include velocity, biases and pose.

The initialization module proceeds by first constructing many intermediate TSMs between adjacent frames and then utilizing this information for high-frequency event-based vision-only localization. This procedure will give initial poses for each frame, and requires the solution 
of individual optimization problems that minimize the values of the negated TSM field at the reprojected semi-dense pixel locations for each frame. The problem formulation for an individual frame is given by
\begin{align}
\label{equa:event_factor}
    \mathop{\arg\min}\limits_{\boldsymbol{\theta}, \mathbf{p}} \sum_{P_j \in \mathcal{D}} \rho(\overline{\mathcal{T}}(\pi_e(\mathbf{R}^c_b \mathbf{R}^T(\boldsymbol{\theta})(\mathbf{P}_j - \mathbf{p}) + \mathbf{t}^c_b))),
\end{align}
where $\rho$ is a robust loss function (e.g. Huber loss), $\pi_e$ is a function which projects the 3D points in event camera frame onto the image plane, $\mathbf{R}(\cdot)$ is the function that maps a minimal rotation representation to a 3-by-3 rotation matrix, and $\mathbf{P}_j \in \mathcal{D}$ is a 3D world point within the current field of view of the event camera.

%High frequency intermediate poses would be propagated into keyframe poses.
When sufficiently many frames for bootstrapping have been collected, we adopt the IMU measurement residual function from \cite{qin2018vins} and loosely align the IMU pre-integration terms with the event-based vision-only pose estimations.
The IMU measurement residual function is given by
\begin{equation}
\label{equa:imu_factor}
\renewcommand*{\arraystretch}{1.5}
\scalebox{0.8}{
$
\mathbf{r}( \boldsymbol{\mathcal{S}}_i, \boldsymbol{\mathcal{S}}_{i+1}) = 
\begin{bmatrix}
\mathbf{R}^T(\boldsymbol{\theta}_i)(\mathbf{p}_{i+1} - \mathbf{p}_{i} + \frac{1}{2}\mathbf{g}^w\Delta t_i^2 - \mathbf{v}_i\Delta t_i) - \hat{\boldsymbol{\mathcal{\alpha}}}^i_{i+1}  \\
\mathbf{R}^T(\boldsymbol{\theta}_i)(\mathbf{v}_{i+1} + \mathbf{g}^w\Delta t_i - \mathbf{v}_i) - \hat{\boldsymbol{\beta}}^i_{i+1} \\
2[\mathbf{Q}(\boldsymbol{\theta}_i)^{-1} \otimes \mathbf{Q}(\boldsymbol{\theta}_{i+1}) \otimes (\hat{\boldsymbol{\gamma}}^i_{i+1})^{-1}]_{xyz} \\
\mathbf{b}_{\alpha_{i+1}} - \mathbf{b}_{\alpha_{i}} \\
\mathbf{b}_{\omega_{i+1}} - \mathbf{b}_{\omega_{i}}
\end{bmatrix}
$
},
\end{equation}
where $\mathbf{Q}(\cdot)$ transforms the minimal rotation representation to a quaternion and $\hat{\boldsymbol{\mathcal{\alpha}}}^i_{i+1}, \hat{\boldsymbol{\beta}}^i_{i+1}, \hat{\boldsymbol{\gamma}}^i_{i+1}$ represent the IMU pre-integration vector using raw IMU measurements.
Given $N$ frames, the loosely coupled visual-inertial alignment objective is given by
\begin{align}
     \mathop{\arg\min}\limits_{\mathbf{v}_i,\mathbf{b}_{\alpha_i}, \mathbf{b}_{\omega_i}} \sum_{i=1}^{N-1} ||\mathbf{r}( \boldsymbol{\mathcal{S}}_i, \boldsymbol{\mathcal{S}}_{i+1})||^2_{\boldsymbol{\Sigma}{i, i+1}}.
\end{align}
Note that we fix the keyframe poses and optimize the remaining states only. Since the existing IMU pre-integrations are based on the old states, they should be re-propagated using the optimized state once the initialization is finished.
% Once the optimization has converged, the IMU pre-integration terms are re-propagated using the new biases.

\subsection{Sliding window optimization}
\label{subsec:optimization}
The initialization module constructs a window of frames with initial states. Once given, we use a second-order motion model to obtain predictions for each new keyframe state. In practice, the field of view of frames within the sliding window typically only covers part of the semi-dense map. In order to reduce the complexity of the optimization, we maintain a set of activate points. The set will keep the points visible in the last sliding window optimization. We also construct a wider frustum for new keyframes using their predicted pose to select potentially visible new points and merge them into the activate points set.
% In order to further utilize the relative information brought by IMU measurements and reduce the complexity of the optimization, we only maintain points that are observed by at least two frames as potential candidate points for optimization. 
We randomly sample a fixed-size subset $\mathcal{P}$ from this set of points to ensure the optimization maintain approximately constant time complexity.

Fig.~\ref{fig:factor_graph} illustrates the factor graph of the concluding tightly-coupled sliding window optimization framework. The optimization is executed each time a new frame is added to the window and the oldest frame is removed. It is furthermore important to remain consistent with mature nodes that already left the sliding window such that a smoother trajectory can be maintained. Here we choose a simple solution that consists of fixing the first node in the graph. Combining the residuals in \eqref{equa:imu_factor} and \eqref{equa:event_factor}, the objective of the event-based visual-inertial sliding window optimizer finally becomes
\begin{align}
    \mathop{\arg\min}\limits_{\boldsymbol{\mathcal{S}}_1, \dots, \boldsymbol{\mathcal{S}}_N} \sum_{\mathbf{P}_j \in \mathcal{P}}&\sum_{i = 1}^N \rho(\overline{\mathcal{T}}(\pi_e(\mathbf{R}^c_b \mathbf{R}^T(\boldsymbol{\theta}_i)(\mathbf{P}_j - \mathbf{p}_i) + \mathbf{t}^c_b))) \notag \\
    &+  \sum_{i = 1}^{N-1} ||\mathbf{r}( \boldsymbol{\mathcal{S}}_i, \boldsymbol{\mathcal{S}}_{i+1})||^2_{\boldsymbol{\Sigma}_{i, i+1}}.
\end{align}
where $N$ is the sliding window size and $\mathcal{P}$ is the maintained activate point set.
We use Ceres solver \cite{Agarwal_Ceres_Solver_2022} to solve all above nonlinear problems.

\section{EXPERIMENTS}

We now proceed to the experimental evaluation of our event-based visual-inertial tracking framework. All experiments are conducted on the \textit{small scale} sequences of the publicly available \textbf{VECtor} dataset \cite{gao2022vector}. The dataset contains event stereo cameras (Prophesee Gen3 CD) with VGA resolution ($640 \times 480$), only the left one of which we use, and a nine-axis IMU (XSens MTi-30 AHRS) at 200Hz. The ground truth is generated by the OptiTrack motion capture system. All sensors are well synchronised though a micro-controller unit at hardware level. We noticed that at certain moments in some sequences, there is only a small overlap between the event camera's field of view and the reconstructed semi-dense map, which leads to tracking failures for all methods. Therefore, we apply clipping as necessary on certain sequences. We first compare the proposed \textbf{EVIT} framework against several state-of-art event-based localization approaches on both normal and challenging sequences. To conclude, we conduct an ablation study using different motion models to analyze the impact of using IMU predictions on tracking performance. We provide both qualitative and quantitative results to show the robustness and effectiveness of our method.

We use the \textit{evo} odometry evaluation tool \cite{grupp2017evo} to align trajectories and conduct comparisons based on the RMSE of the average trajectory error (ATE) metric. The trajectories are aligned by an SE(3) transformation derived from the first pair of the synchronized camera and ground truth poses.

% We show that introducing higher-order motion information into the system enhances robustness, accuracy and efficiency compared to the event vision-only tracking system.

\subsection{Evaluation on VECtor}

\begin{figure*}[t]
	\centering
	\subfigure[translation estimates] {
         \centering
        \includegraphics[width=0.99\columnwidth]{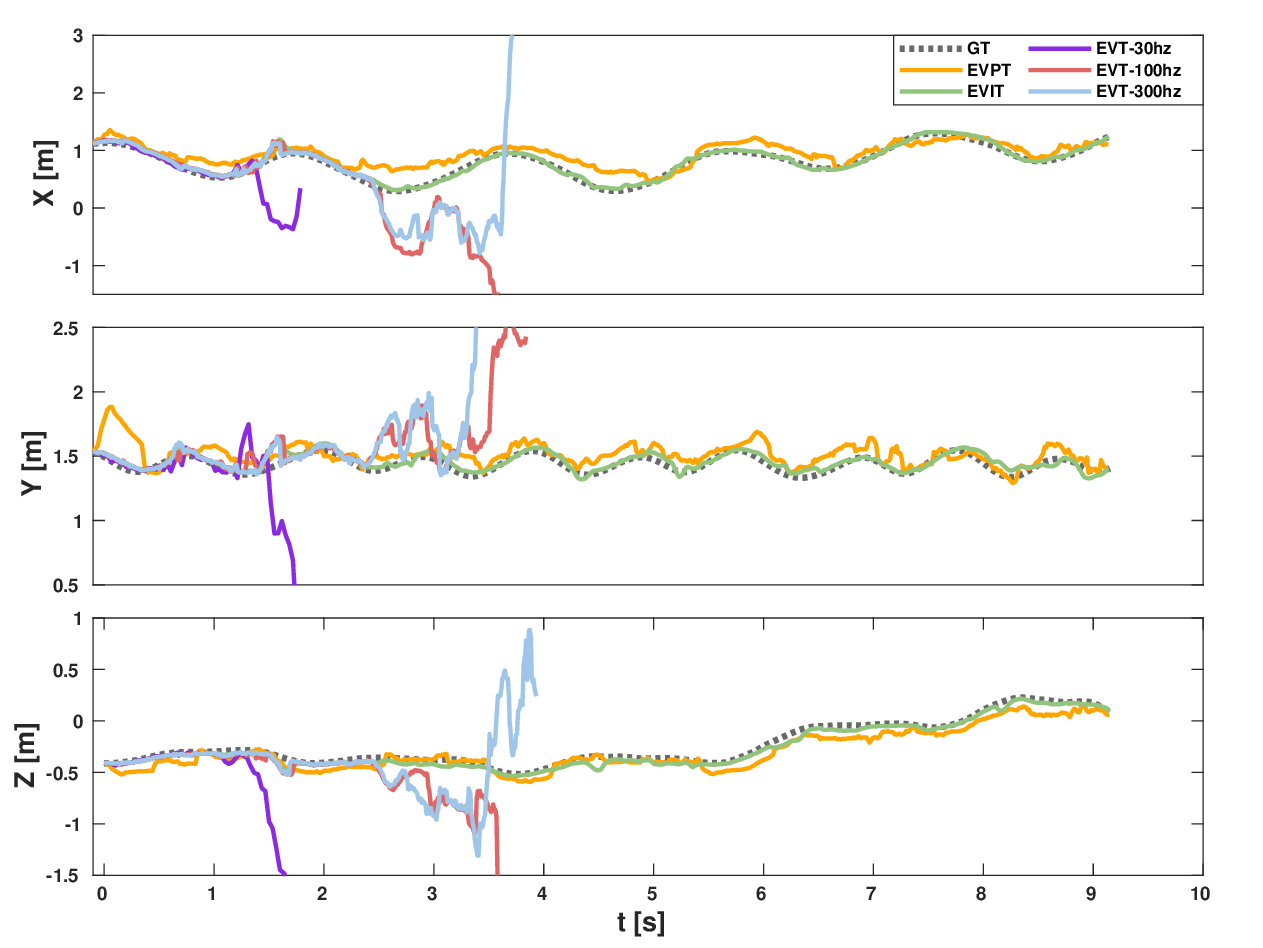}
        }
	\subfigure[rotation estimates] {
         \centering
        \includegraphics[width=0.99\columnwidth]{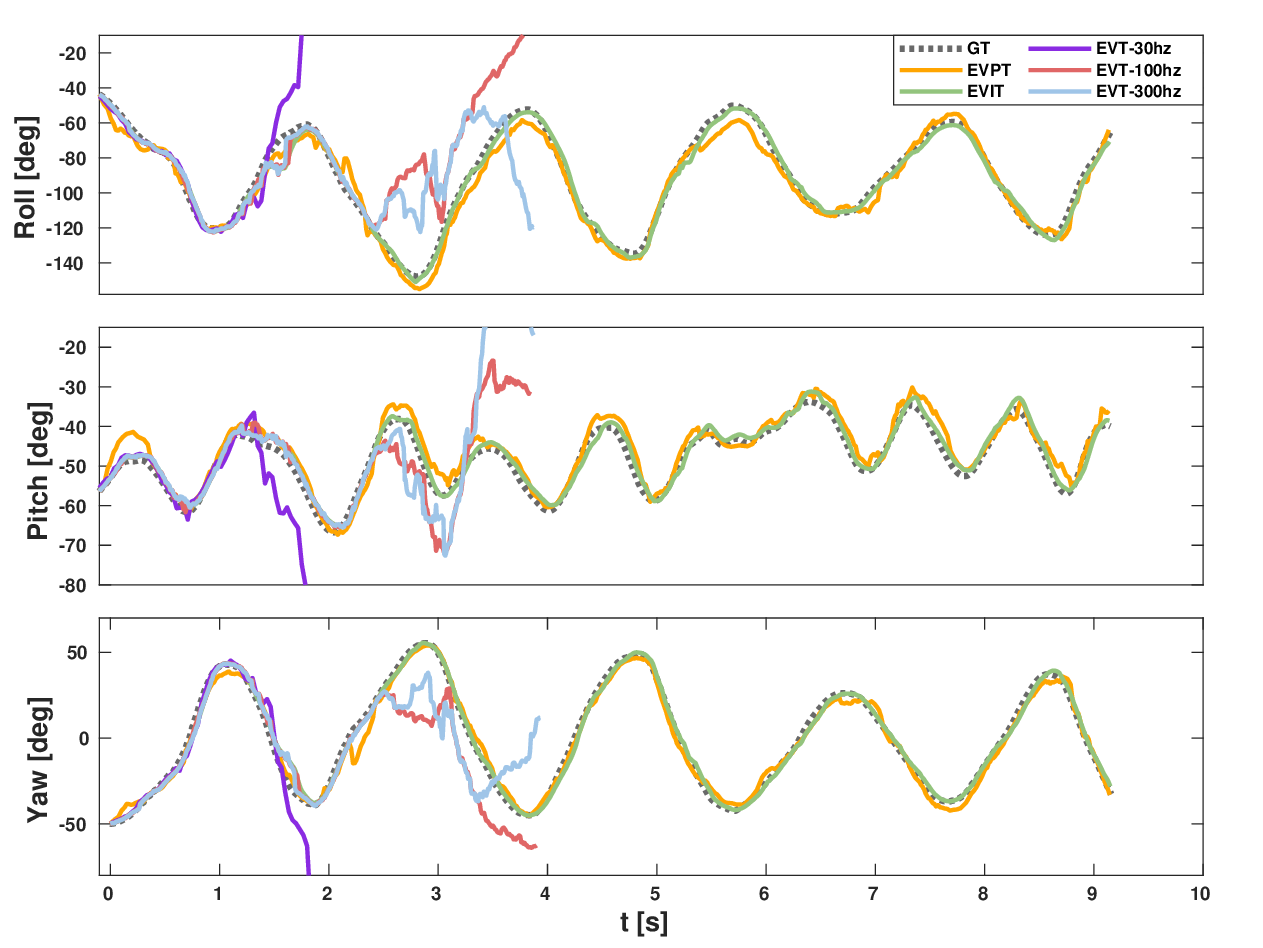}
        }
	\caption{Comparison of trajectories generated by various pure and inertial-supported semi-dense event-based tracking solutions, as well as a photometric alternative. The sequence is \textit{sofa\_fast} from the \textbf{VECtor} benchmark~\cite{gao2022vector}. The method proposed in this work is called \textbf{EVIT}.}
	\label{fig:evo_traj}
\end{figure*}
In the first experiment, we compare the performance of the proposed system with existing event-based localization frameworks. Note that the present work is an extension of the event vision tracking module Canny-EVT presented by Zuo et al.~\cite{zuo2024cross}. As mentioned in there, the \textbf{EVT} module relies on a high-frequency localization method which trades efficiency for robustness, thereby resulting in an offline method with no real-time capabilities. We compare our proposed \textbf{EVIT} against \textbf{EVT} running at different rates and on the same semi-dense map.

The global semi-dense map is constructed from the open-source framework \cite{he2018incremental}, which uses the probabilistic semi-dense mapping technique proposed by Mur-Artal and Tardos \cite{mur2015probabilistic}. Note that the constructed edges do not necessarily need to be geometric edges in 3D and may still be appearance edges. In order to avoid the scale ambiguity issue caused by monocular slam, we replace each keyframe pose by the corresponding ground truth pose. We also manually transform the reconstructed semi-dense map to a global world reference frame where the negative $z$-axis direction aligns with the direction of gravity to ensure consistency with IMU data.

\begin{table}[htbp]
\caption{Absolute trajectory error (ATE) on normal sequences. Position: {[}cm{]}, Orientation: {[}$^{\circ}${]} }
\label{tab:normal}
\resizebox{\columnwidth}{!}{%
\begin{tabular}{@{}ccccccccc@{}}
\toprule
\multirow{2}{*}{Sequence} &
  \multirow{2}{*}{Method} &
  \multirow{2}{*}{Frequency} &
  \multicolumn{2}{c}{30\% sequence} &
  \multicolumn{2}{c}{50\% sequence} &
  \multicolumn{2}{c}{100\% sequence} \\ \cmidrule(l){4-9} 
                              &                      &       & Pos.          & Orient.       & Pos.          & Orient.       & Pos.          & Orient.       \\ \midrule
\multirow{5}{*}{sofa\_normal} & \multirow{3}{*}{EVT} & 300hz & 3.26          & 1.34          & 3.66          & 1.75          & 3.37          & 1.65          \\
                              &                      & 100hz & \textbf{3.16} & \textbf{1.33} & 6.98          & 2.70          & 5.43          & 2.28          \\
                              &                      & 30hz  & 3.54          & 1.58          & -             & -             & -             & -             \\
                              & EVPT                 & (dyn)   & 15.64         & 5.08          & 15.27         & 5.50          & -             & -             \\
                              & EVIT                 & (dyn)   & 3.23          & 1.40          & \textbf{3.23} & \textbf{1.56} & \textbf{3.15} & \textbf{1.53} \\ \midrule
\multirow{5}{*}{robot\_normal} &
  \multirow{3}{*}{EVT} &
  300hz &
  0.91 &
  \textbf{0.89} &
  \textbf{1.04} &
  \textbf{1.08} &
  1.09 &
  1.31 \\
                              &                      & 100hz & 0.97          & 0.91          & 1.09          & 1.09          & 1.16          & 1.31          \\
                              &                      & 30hz  & 1.02          & 0.90          & 1.11          & 1.11          & 1.20          & 1.35          \\
                              & EVPT                 & (dyn)  & 15.36         & 9.927         & 13.40         & 8.39          & 14.17         & 9.03          \\
                              & EVIT                 & (dyn)  & \textbf{0.84} & \textbf{0.89} & \textbf{1.04} & 1.10          & \textbf{1.00}          & \textbf{1.14}          \\ \midrule
\multirow{5}{*}{desk\_normal} &
  \multirow{3}{*}{EVT} &
  300hz &
  \textbf{1.21} &
  \textbf{0.74} &
  1.87 &
  \textbf{0.81} &
  2.25 &
  \textbf{0.88} \\
                              &                      & 100hz & 1.34          & 0.77          & 1.87          & 0.84          & \textbf{2.14} & 0.92          \\
                              &                      & 30hz  & \textbf{1.21} & 0.77          & 1.92          & 0.87          & 2.30          & 0.96          \\
                              & EVPT                 &  (dyn) & 7.71          & 1.83          & 11.02         & 2.61          & 16.08         & 3.33          \\
                              & EVIT                 &   (dyn) & 1.32          & 0.80          & \textbf{1.79} & 0.87          & 2.22          & \textbf{0.94} \\ \bottomrule
\end{tabular}%
}
\end{table}

We also compare the \textbf{EVIT} with the photometric event-based camera localization method proposed by Bryner et al.~\cite{Bryner19icra}. We refer to it here as \textbf{EVPT}. The photometric map is directly constructed using the RGBD data and ground truth poses contained in the dataset. We directly re-project all valid pixels back into the 3D space and carefully polish the map by hand to reduce edge effects caused by the RGBD observations.

Note that not all methods can track the entire sequence so we define three milestones at 30\%, 50\% and 100\% of the sequence. The corresponding error is simply not displayed if a certain milestone is not reached. Since \textbf{EVIT} and \textbf{EVPT} all adopt dynamic frame determination schemes, the rate depends on the motion and the scene. The actual rate is lower for mild motion and higher during aggressive motion, and typically valies between 30 and 60 Hz. As shown in  Table.~\ref{tab:normal}, though the improvement brought by \textbf{EVIT} is not significant on normal sequences, both \textbf{EVT} and \textbf{EVIT} outperform \textbf{EVPT}. This is sensible given that photometric localization methods strongly rely on the consistency of the actual light conditions when the map was built, which is not only hard to achieve for varying times of the day, but also hard to achieve across different sensors. Furthermore, photometric methods involve error metrics with no direct geometric meaning, and it is thus hard to ensure optimality of the motion estimation. In terms of EVIT, it can be observed that the addition of the IMU arguably offers only a limited improvement over the event-only alternative when the motion is normal and not exhibiting strong dynamics. However, EVIT generally produces small errors, often leading to the absolutely best tracking result. 

% \begin{table}[b!]
\begin{table}[htbp]
\caption{Absolute trajectory error (ATE) on challenging sequences. Position: {[}cm{]}, Orientation: {[}$^{\circ}${]} }
\label{tab:challenging}
\resizebox{\columnwidth}{!}{%
\begin{tabular}{@{}ccccccccc@{}}
\toprule
\multirow{2}{*}{Sequence} &
  \multirow{2}{*}{Method} &
  \multirow{2}{*}{Frequency} &
  \multicolumn{2}{c}{30\% sequence} &
  \multicolumn{2}{c}{50\% sequence} &
  \multicolumn{2}{c}{100\% sequence} \\ \cmidrule(l){4-9} 
                             &                      &       & Pos.          & Orient.       & Pos.          & Orient.       & Pos.          & Orient.       \\ \midrule
\multirow{5}{*}{sofa\_fast}  & \multirow{3}{*}{EVT} & 300hz & 20.45         & 7.84          & -             & -             & -             & -             \\
                             &                      & 100hz & 22.66         & 8.56          & -             & -             & -             & -             \\
                             &                      & 30hz  & -             & -             & -             & -             & -             & -             \\
                             & EVPT                 & (dyn)      & 25.29         & 6.53          & 26.32         & 6.77          & 23.24         & 6.18          \\
                             & EVIT                 & (dyn)      & \textbf{9.02} & \textbf{4.02} & \textbf{8.11} & \textbf{3.60} & \textbf{7.08} & \textbf{3.22} \\ \midrule
\multirow{5}{*}{robot\_fast} & \multirow{3}{*}{EVT} & 300hz & 3.76          & 2.83          & 4.12          & 2.92          & -             & -             \\
                             &                      & 100hz & 5.47          & 3.67          & 9.15          & 4.49          & -             & -             \\
                             &                      & 30hz  & -             & -             & -             & -             & -             & -             \\
                             & EVPT                 & (dyn)      & -             & -             & -             & -             & -             & -             \\
                             & EVIT                 & (dyn)      & \textbf{3.22} & \textbf{2.67} & \textbf{3.24} & \textbf{2.80} & \textbf{3.85} & \textbf{2.98} \\ \midrule
\multirow{5}{*}{desk\_fast}  & \multirow{3}{*}{EVT} & 300hz & 4.03          & 2.47          & 4.52          & 2.35          & 14.65         & 7.95          \\
                             &                      & 100hz & 2.24          & 2.19          & 2.48          & 2.52          & -             & -             \\
                             &                      & 30hz  & 5.13          & 3.35          & 15.4          & 7.23          & -             & -             \\
                             & EVPT                 & (dyn) & 5.80          & 2.91          & 5.63          & 3.99          & 7.13          & 4.27          \\
                             & EVIT                 & (dyn) & \textbf{1.53} & \textbf{1.69} & \textbf{1.61} & \textbf{2.23} & \textbf{3.59} & \textbf{3.01} \\ \bottomrule
\end{tabular}%
}
\end{table}

In contrast, the numerical results shown in Table.~\ref{tab:challenging} illustrate a substantial improvement in accuracy and robustness in challenging scenarios. The majority of the alternative methods either fail to complete the entire sequence or result in large localization errors. Fig.~\ref{fig:evo_traj} visualizes a comparison of actual trajectories from one of the more challenging sequences, demonstrating how all semi-dense comparison methods that do not employ inertial readings fail after at most 4s. 

\subsection{Ablation study on motion model}
The tracking process in this work is based on the nonlinear optimization technique which is often sensitive to initial state given by motion model. 
Therefore, we also conduct an ablation study to evaluate the effect of different motion models. The latter can be adopted based on the system's available sensors. When there is no internal motion information, a simple strategy would be the initialize the pose of a newly incoming keyframe with the pose of the previous one, a paradigm we may refer to as a zeroth-order motion model. A more advanced model could be the first-order motion model (i.e. constant velocity model), which assumes that the velocity remains constant over small periods of time, and thereby permits the initialization of a new keyframe with a relative pose to the previous one that is similar to the relative pose between the previous keyframe and its preceding one. Equipped with an IMU, we can introduce the second-order motion model, which uses acceleration and angular velocity readings to propagate the previous pose to the current one. 

\begin{figure}[htbp]
	\centering
	\subfigure[zeroth-order model] {\includegraphics[width=.15\textwidth]{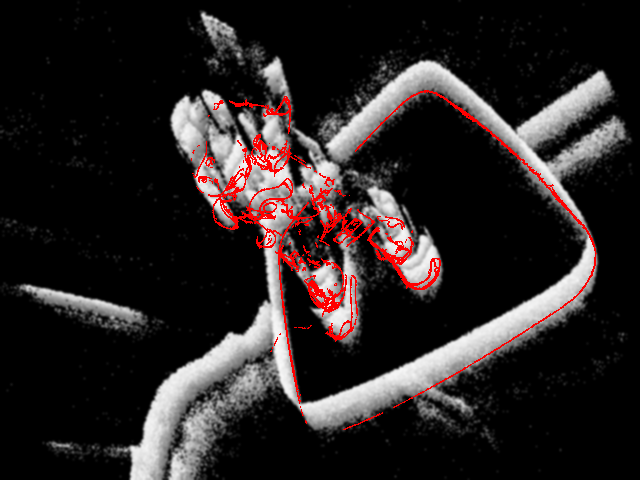}}
	\subfigure[first-order model] {\includegraphics[width=.15\textwidth]{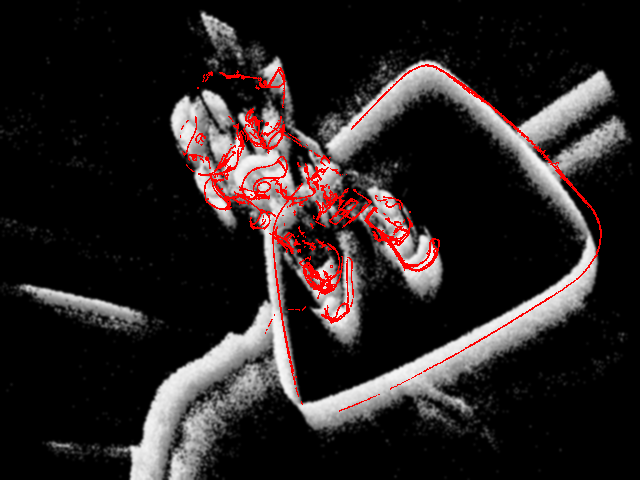}}
	\subfigure[second-order model] {\includegraphics[width=.15\textwidth]{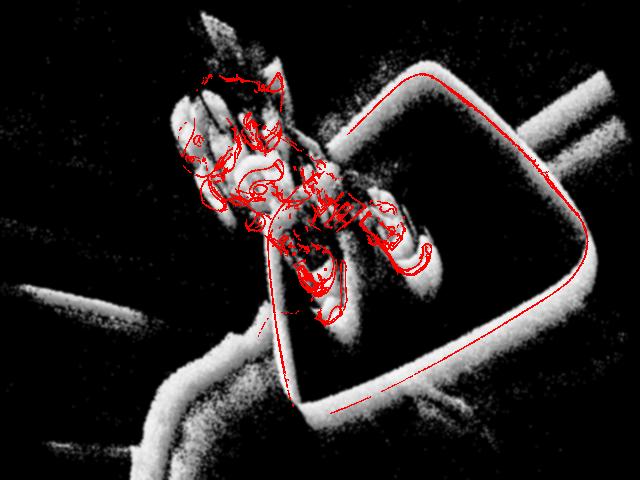}}
	\caption{Initial projection of semi-dense cloud on \textbf{TSM} using different motion models. }
	\label{motion_model}
\end{figure}

The three motion models are tested with the same tightly coupled optimization back end. Fig.~\ref{motion_model} shows qualitative results for the different models, demonstrating how the second-order motion model provides better initial poses for the optimization back end. The full ATE results are listed in Table.~\ref{ate_motion_model}. Although the differences are not significant on mild sequences, it shows that even under the same optimization scheme, the motion model has a crucial implication in fast-motion scenarios.
\begin{table}[htb]
\caption{Absolute trajectory error (ATE) on different motion models. Position: {[}cm{]}, Orientation: {[}$^{\circ}${]}}
\label{ate_motion_model}
\resizebox{\columnwidth}{!}{%
\begin{tabular}{@{}ccccccc@{}}
\toprule
\multirow{2}{*}{Sequence} & \multicolumn{2}{c}{zeroth-order} & \multicolumn{2}{c}{first-order} & \multicolumn{2}{c}{second-order} \\ \cmidrule(l){2-7} 
                           & Pos.          & Orient.          & Pos.              & Orient.     & Pos.            & Orient.        \\ \midrule
sofa\_fast                 & -             & -                & -                 & -           & \textbf{7.08}   & \textbf{3.22}  \\ \midrule
sofa\_normal               & -             & -                & \textbf{2.84}     & 1.60        & 3.15            & \textbf{1.53}  \\ \midrule
robot\_fast                & -             & -                & -                 & -           & \textbf{3.85}   & \textbf{2.98}  \\ \midrule
robot\_normal              & 1.08          & 1.32             & 1.46              & 1.23        & \textbf{1.00}   & \textbf{1.14}  \\ \midrule
desk\_fast                 & -             & -                & -                 & -           & \textbf{3.59}   & \textbf{3.01}  \\ \midrule
desk\_normal               & 3.75          & 1.59             & 2.77              & 1.16        & \textbf{2.22}   & 0.94  \\ \bottomrule
\end{tabular}%
}
\end{table}

\section{CONCLUSION}

In this work, we have introduced a novel event-inertial tracking framework that aligns with previous approaches in that it proposes the registration of a semi-dense point cloud with contours observed in the time surface map. However, rather than simply aligning individual frames, the introduction of inertial signals permits the joint alignment of multiple adjacent keyframes through the addition of pair-wise regularization terms. The proposed sliding window alignment shows benefits in tracking robustness as the contours in a single TSM are often orthogonal to the projected instantaneous direction of motion, and thus do not fully constrain an individual pose. Furthermore, the addition of the IMU leads to better initializations and thereby improves tracking performance in highly dynamic situations without a substantial increase in frame rate. Interestingly, the proposed windowed tracking method could be equally applied to regular cameras. 

\section{Acknowledgement}
We would like to acknowledge the funding support provided by project 62250610225 by the Natural Science Foundation of China, as well as projects 22DZ1201900, 22ZR1441300, and dfycbj-1 by the Natural Science Foundation of Shanghai.



% \section*{ACKNOWLEDGMENT}

{
    \small
    \bibliographystyle{IEEEtran}
    \bibliography{IEEEexample}
}

\end{document}